# Leveraging Foundation Models for Clinical Text Analysis

Shaina Raza
*Dalla Lana School of Public Health*
*University of Toronto*
Toronto, Canada
shaina.raza@utoronto.ca

Syed Raza Bashir
*Department of Computer Science*
*Toronto Metropolitan University*
Toronto, Canada
syedraza.bashir@torontomu.ca

*Abstract*— **Infectious diseases are a significant public health concern globally, and extracting relevant information from scientific literature can facilitate the development of effective prevention and treatment strategies. However, the large amount of clinical data available presents a challenge for information extraction. To address this challenge, this study proposes a natural language processing (NLP) framework that uses a pre-trained transformer model fine-tuned on task-specific data to extract key information related to infectious diseases from free-text clinical data. The proposed framework includes three components: a data layer for preparing datasets from clinical texts, a foundation model layer for entity extraction, and an assessment layer for performance analysis. The results of the evaluation indicate that the proposed method outperforms standard methods, and leveraging prior knowledge through the pre-trained transformer model makes it useful for investigating other infectious diseases in the future.**

## I. INTRODUCTION

Extracting entities from patient health records data through natural language processing (NLP) techniques can provide valuable insights for researchers, healthcare providers, and public health officials in their efforts to better understand and combat COVID-19. The extraction process involves identifying and extracting specific pieces of information such as patient demographics, symptoms, lab results, and treatment outcomes. These entities can then be used to identify trends and patterns, identify at-risk populations, monitor the spread of the virus, improve patient outcomes, and automate data collection.

Using foundation models [1] in conjunction with NLP techniques is a promising approach for improving the efficiency and accuracy of entity extraction from patient health records. These pre-trained models have been trained on large datasets to learn general-purpose representations of data, which can be fine-tuned on specific task-specific datasets to achieve good performance on the target task. There are several pre-trained models [2] available such as BERT, GPT-2, and RoBERTa that can be used as foundation models, depending on the specific domain and task at hand. This approach has the potential to provide valuable insights and aid in the fight against the pandemic (e.g., COVID-19 [3]), and is a motivation for ongoing research in the field. The specific objectives of the research are:

- To use pre-trained foundation models to extract relevant information such as disease disorders, symptoms, conditions, demographics, lab results, and treatment outcomes from patient health records.

- To fine-tune the foundation model on a task-specific dataset prepared from scientific publications in order to improve the efficiency and accuracy of entity extraction in the absence of patient health records data.

- To gain a better understanding of the disease and develop effective interventions to control its spread and improve patient outcomes by leveraging the insights gained from the extracted entities.

Our research proposes an NLP framework comprising of three main layers: data, foundation model, and assessment. The data layer collects and preprocesses scientific texts from biomedical publications to create a dataset. The foundation model layer employs a pre-trained foundation model, which adapts to the task of extracting relevant information from clinical texts. The assessment layer provides insight into the accuracy and efficiency of the framework. Our proposed framework effectively integrates different NLP models to extract valuable information from clinical texts, which is our main contribution. Particularly, the use of a foundation model for entity extraction offers high effectiveness at capturing language patterns and relationships, robustness to noisy and unstructured text data, and better performance compared to traditional ML approaches. The proposed approach is generalizable, but our focus in this paper is on COVID-19.

## II. LITERATURE REVIEW

Named entity recognition (NER) [4] is a subtask of NLP that involves identifying and classifying named entities in text into predefined categories such as person names, organizations, locations, medical codes, etc. Biomedical NER is a specialized NER task that focuses on identifying and classifying biomedical entities, such as genes, proteins, and diseases, in unstructured text [5]. State-of-the-art biomedical NER models include BiLSTM-based [6] and Transformer-based models [7]–[10], which can capture contextual dependencies and are robust to noise and variations in the input data.

Several works focus on extracting clinical information, particularly related to COVID-19, from unstructured text data. For example, some propose neural event extraction frameworks using BiLSTM-CRF [11] or Transformer architectures to identify and classify diagnoses, symptoms, and other clinical events [10]. Other works present NER models based on BiLSTM-CNN architectures for extracting symptoms from unstructured COVID-19 data [12]. These works have the potential to be used for tasks such as public health surveillance and monitoring.

Some NER methods also identify non-clinical factors like social determinants of health (SDOH) in addition to a variety of clinical factors [13], [14]. This is significant because SDOH factors have a significant impact on health outcomes, particularly during a pandemic like COVID-19.

In this work, we extract the key information (clinical and SDOH) from scientific literature about infectious diseases using NLP techniques. In particular, we include a foundation model [15] layer for extracting entities from texts. Foundation models are pre-trained deep learning models that have been trained on vast amounts of data to capture general patterns and relationships within the data [15].

## III. METHODOLOGY

The proposed framework, shown in Figure 1, consists of data, foundation model and assessment layers, which are explained next.

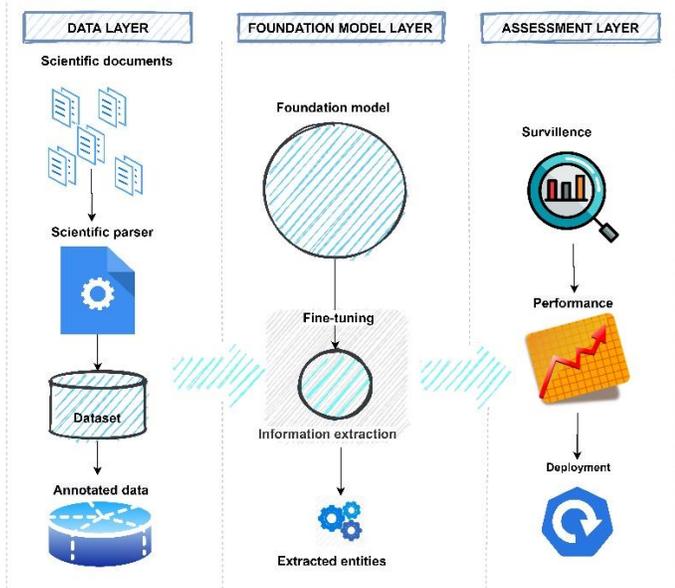

Fig. 1. Proposed NLP framework

### A. Data layer

In our NLP framework, we collect and process scientific texts from publications using the LitCOVID API [16]. The data layer includes a scientific parser that extracts texts from the publication files and preprocesses them by concatenating the texts to create a dataset with columns for paper ID and text. To annotate a portion of the data (500 case report publications), we use the Spark JohnSnowLabs [17] annotation lab and employ active learning [18] for corpus re-annotation. The named entities used in our work are inferred from the pre-trained NER model, NER_JSL [19] from JohnSnowLabs. To measure inter-annotator agreement [20], we use the simple agreement method, which calculates the percentage of annotations that match across all annotators without taking random chance into account. These annotations are saved in the CONLL-2003 [21] format, which is a prototypical data format for training NER tasks. In total, we obtain approximately 1500 sentences with gold labels at the end of this step. Appendix A provides a list of named entities used in our work.

### B. Foundation model layer

A foundation model is a pre-trained deep learning model that serves as the starting point for training a custom model [1] . These models are trained on large amounts of text data and can be fine-tuned on a smaller dataset specific to the task at hand, which can improve performance and reduce the required training data.

In this work, we use BioBERT [22] as our foundation model for extracting biomedical named entities from scientific texts. BioBERT is a transformer-based model that has been pre-trained on a large clinical dataset and has shown strong performance in understanding the context and meaning of words within a sentence.

To adapt the model to our specific task, we fine-tune the pre-trained BioBERT on our annotated data. This fine-tuning process helps the model learn the naming conventions and entities that are relevant for the task, and enables the model to perform tasks such as entity extraction, linking, and classification.

For the NER task, a foundation model like BioBERT is essential for fine-tuning in the biomedical domain. The BioBERT model consists of several layers, including an embedding layer that converts input tokens into continuous vectors, a multi-layer transformer encoder that learns representations of the input text, a pooler layer that generates a fixed-length representation of the input text, an NER layer that is fine-tuned for the specific task, and an output layer that generates a probability distribution over possible entities for each token in the input text.

### C. Assessment layer

The assessment or evaluation layer is a crucial component of the proposed NLP-based framework for extracting data from scientific texts. This layer is responsible for measuring the performance of the fine-tuned model on a specific task such as NER by using different performance metrics. These metrics can include measures such as precision, recall, and F1-score, which are commonly used to evaluate the performance of NER models.

This layer also provides insights into the effectiveness of the model in the surveillance task of public health. This is achieved by evaluating the model ability to identify and extract relevant information from texts. For example, the model ability to accurately identify and extract named entities such as diseases, symptoms, and treatments can be evaluated. The performance on this task would give an insight on how effective the model is in extracting relevant information for public health surveillance.

Additionally, this layer also provides an understanding of the limitations of the model by identifying the errors made by the model, which can be used to further improve the model. By evaluating the model on a specific task, this layer helps to ensure that the model is performing well and is suitable for use in real-world applications.

By integrating all these layers, we present a framework that has the potential to enhance the efficiency and accuracy of public health surveillance by automating the process of extracting relevant information from scientific texts.

## IV. EXPERIMENT AND RESULTS

### A. Data

This study uses the LitCOVID API [23] to gather clinical publications (e.g., case reports) related to COVID-19 patients. The criterion for inclusion is that the publications must be in English and exclude grey literature, preprints, and clinical trials. The study focuses on patients aged 12 or older and the data collection period is defined as between March 1st, 2022, and June 30th, 2022. The total number of publications in the dataset, after performing may filtrations, is approximately 15k.

### B. Experimental Settings

In the conducted experiment, several hyperparameters were tuned to obtain the best performance for the proposed NER model. The dropout rate was set to 0.5 with a range of 0.2 to 0.7, while the LSTM state size was set to 200 with a range of 200 to 300. The batch size was set to 16 with a range of 8 to 128, and the number of epochs was set to 40 with a range of 20 to 80. The learning rate was set to 1.e-05 with a range of 1.e-9 to 1.e-2, and the lr decay coefficient was set to 0.005 with a range of 0.001 to 0.01. The warmup steps were set to 10,000, and the optimizer used was ADAM with $\beta_1=0.9$ and $\beta_2=0.999$. The word dimension was set to 300 with a range of 50 to 450, and the hidden size of the LSTM (baseline) was set to 300 with a gradient clipping of 5.0.

The fine-tuning transformer-based architectures (BioBERT and others) had a maximum sequence length of 128, with 12 layers, 12 attention heads, and an embedding size of 768. For different datasets, the fine-tuning took different hours, ranging from 2 to 10 hours for our dataset. In the NER task, the length of sentences was fixed to 512.

Furthermore, the performance of the proposed NER model was evaluated using different training and testing sizes. A random selection of 70% and 30% of annotated data was used for training and testing, respectively, for the proposed model. For performance comparison, the test sets from the NCBI-disease [24], i2b2-clinical [25] and i2b2-2012 [26] benchmark datasets were used along with our test set. The micro-average F1-score was used to evaluate the performance of NER tasks, as it takes into account all the true positives, false positives, and false negatives, providing a single score that reflects the overall performance of the model. The BILSTM-CNN-CHAR [27] and BERT [28] models were used as baselines and optimized to their optimal settings, and the best results for each method were reported. The BERT encoder layers are implemented using the Huggingface library.

### C. Overall Comparison

In this experiment, we compare the performance of our NER model in the foundation layer by comparing its performance with baseline methods. The results of the evaluation are presented in Table I and discussed next.

TABLE I. COMPARISON WITH BASELINES.

| Model/Dataset | NCBI | i2b2-clinical | i2b2-2012 | Our test set |
|---|---|---|---|---|
| BILSTM-CNN-Char | 88.73 | 84.08 | 83.19 | 89.20 |
| BERT | 88.15 | 88.95 | 87.43 | 92.34 |
| **Our fine-tuned** | **89.83** | **89.12** | **90.13** | **92.78** |

The results, presented in Table 1, indicate that our proposed method outperforms the baselines on all datasets. We found that using BERT-based methods, such as BioBERT, generally yielded better results than BiLSTM-based methods. Specifically, our method achieved a high F1-score of 89.83% on the NCBI-disease dataset. This suggests that the use of BioBERT as a foundation model can significantly improve the performance of NER models without the need for extensive feature engineering.

Additionally, we observed that using clinical embeddings from BioBERT improved the performance on i2b2 datasets. This highlights the importance of using pre-trained models that are specialized for the domain of interest. In this case, the use of BioBERT, which is pre-trained on biomedical data, greatly improved the model ability to recognize entities in clinical text.

Finally, we found that fine-tuning pre-trained models was a key factor in achieving high performance. By adjusting the pre-trained model to the specific task, the model performance was significantly improved. This highlights the potential of using a foundation model for NER task in the biomedical domain.

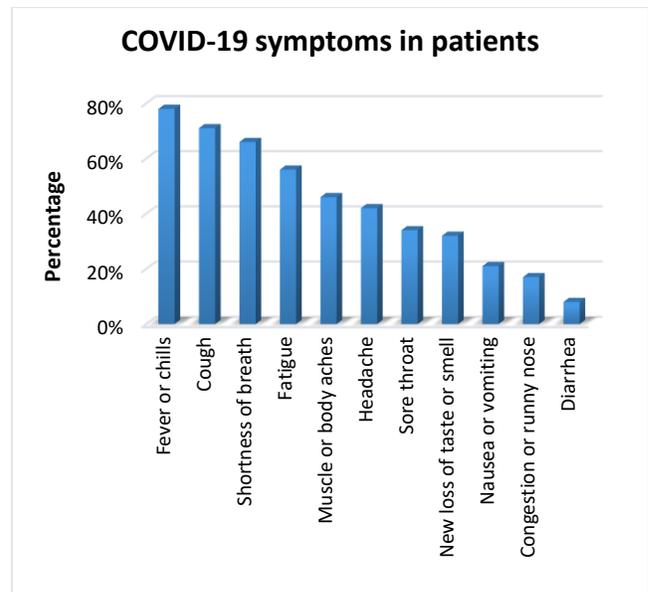

Fig. 2. Percentage of COVID-19 patients with identified selected symptoms reported through proposed NER model.

Identify applicable funding agency here. If none, delete this text box.

Next, we show the model performance in extracting COVID-19 entities. According to results in Figure 2, the symptoms listed (fever or chills, cough, shortness of breath, fatigue, muscle or body aches, headache, sore throat, new loss of taste or smell, nausea or vomiting, congestion or runny nose, diarrhea) are common symptoms of COVID-19. This finding also coincide with the results as reported in the literature [29]. Next, we show common tests conducted on patients for COVID-19 testing.

In Figure 3, we also show medical conditions reported in COVID-19 patients and discover that pneumonia, respiratory disorder, myocarditis, and thrombosis are among the comorbidities reported in COVID-19 patients [30] that we found through our NER method.

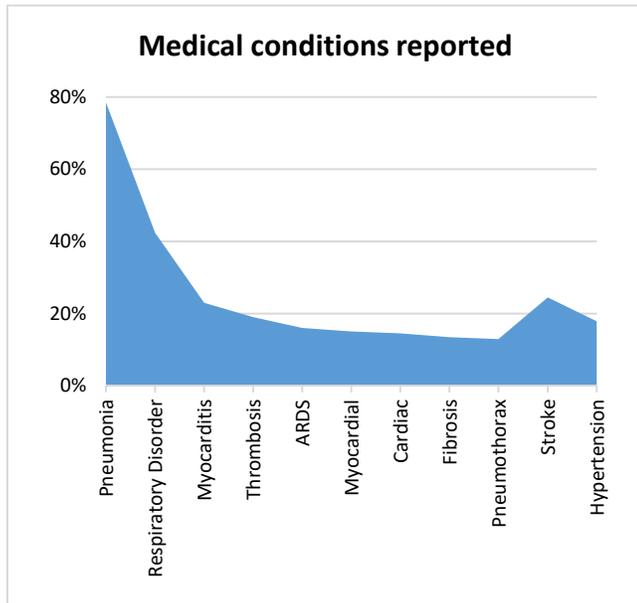

Fig. 3. Percentage of patients with SARS-CoV-2 and other comorbidities.

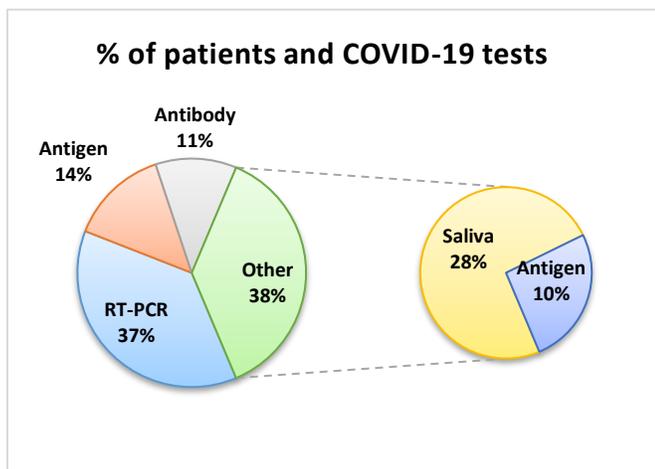

Fig. 4. Percentage of patients with SARS-CoV-2 taking different tests.

Next, we show different COVID-19 tests reported through NER model in Figure 4. The result shows that RT-PCR (Reverse Transcriptase Polymerase Chain Reaction) and antigen tests are commonly used to detect COVID-19 [31], the antibody tests are used to detect if a person has previously been infected with the virus, and saliva can be used as a sample for RT-PCR testing.

Table II shows various diseases and disorders, along with the keywords that are commonly associated with them. The keywords can provide insight into the symptoms, complications, and underlying mechanisms of the diseases.

TABLE II. TOP-3 KEYWORDS RELATED TO MAIN DISEASE DISORDERS REPORTED BY MODEL.

| Disease/Disorder | Keywords Identified |
| --- | --- |
| ARDS | Fluid accumulation, hypertension, lung injury |
| Acute kidney injury | Hyperkalemia, severe metabolic acidosis |
| Blood Clotting | Coagulation, thrombosis, pulmonary embolism |
| Acute respiratory failure | Fibrosis, venous oxygen saturation, shortness of breath |
| Cardiac Injury | Heart failure, inflammation |
| Dry cough | Rhinorrhea, phlegm |
| Hypertension | Blood pressure, renal, cardio |
| COVID-19 | Fever, cough, shortness of breath, loss of smell/taste, pneumonia, cytokine storm |
| Diabetes | Hyperglycemia, insulin resistance, neuropathy, nephropathy |
| Cancer | Tumor, chemotherapy, radiation, metastasis |
| Stroke | Ischemia, hemorrhage, aphasia, hemiplegia |
| Alzheimer's disease | Memory loss, cognitive impairment, neurofibrillary tangles, beta-amyloid plaques |

The results in Table II show that for the ARDS (Acute Respiratory Distress Syndrome) disease disorder, the keywords include fluid accumulation, hypertension, and lung injury. This suggests that ARDS is characterized by respiratory failure and lung damage, often accompanied by high blood pressure and fluid buildup in the lungs.

Similarly, for COVID-19, the keywords include fever, cough, shortness of breath, loss of smell/taste, pneumonia, and cytokine storm. These are the common symptoms and complications associated with COVID-19 infection, which has caused a global pandemic.

Other diseases such as diabetes, cancer, stroke, and Alzheimer's disease also have their unique set of keywords, providing insights into their pathophysiology and clinical features. Overall, these results serve as a useful reference for understanding various diseases and their associated keywords. These conditions can have serious consequences, including organ failure, heart failure, and death, if not treated promptly. It is important to be aware of these potential complications and to seek medical attention if experiencing symptoms.

## V. DISCUSSION AND CONCLUSION

In this study, we proposed a novel NLP-based framework of tools for extracting large amounts of data from scientific texts. The framework is designed to assist with the assessment, diagnosis, and treatment of COVID-19 by analyzing scientific literature and extracting relevant information. The motivation for this work was the COVID-19 pandemic and the need for efficient and accurate data analysis. However, it is important to note that there are a set of factors such as ethics, privacy, and security that should be carefully considered and discussed by researchers when using NLP for healthcare data analysis.

The proposed NLP framework is based on foundation models such as BioBERT, which have been pre-trained on large amounts of text data and have been found to be highly effective at understanding the context and meaning of words within a sentence. The framework uses BioBERT to extract relevant information such as named entities, semantic roles, and relationships between entities from scientific texts. This information can then be used to assist with the assessment, diagnosis, and treatment of COVID-19.

When using a foundation model for biomedical applications, it is important to comply with HIPAA (Health Insurance Portability and Accountability Act) [32] regulations and ensure that patient health information is protected. This includes implementing appropriate measures for data security, such as encryption and access controls, as well as obtaining patient consent for the use of their health information. It is also important to ensure that the dataset used to fine-tune the model does not contain any identifiable patient information.

In future directions, we plan to expand the pipeline by incorporating other data sources such as electronic health records to provide a more comprehensive picture of the outbreak. This will allow us to combine the information extracted from scientific texts with other relevant data sources to gain a better understanding of the disease and its spread. Additionally, we aim to extend the framework to detect other infectious diseases. This will enable us to use the same framework for other outbreaks and pandemics in the future.

We also plan to incorporate explainable AI to make the framework more transparent. This will help to ensure that the decisions made by the model are based on valid and reliable information, and that the framework can be easily understood by researchers, healthcare professionals, and policymakers. Finally, we aim to improve scalability to handle large amounts of data and ensure real-time analysis. This will enable us to analyze large amounts of data in a timely manner, which is crucial for making informed decisions during an outbreak.

APPENDIX A: NAMED ENTITIES.

Social_History_Header, Oncology_Therapy, Blood_Pressure, Respiration, Performance_Status, Family_History_Header, Dosage, Clinical_Dept, Diet, Procedure, HDL, Weight, Admission_Discharge, LDL, Kidney_Disease, Oncological, Route, Imaging_Technique, Puerperium, Overweight, Temperature, Diabetes, Vaccine, Age, Test_Result, Employment, Time, Obesity, EKG_Findings, Pregnancy, Communicable_Disease, BMI, Strength, Tumor_Finding, Section_Header, RelativeDate, ImagingFindings, Death_Entity, Date, Cerebrovascular_Disease, Treatment, Labour_Delivery, Pregnancy_Delivery_Puerperium, Direction, Internal_organ_or_component, Psychological_Condition, Form, Medical_Device, Test, Symptom, Disease_Syndrome_Disorder, Staging, Birth_Entity, Hyperlipidemia, O2_Saturation, Frequency, External_body_part_or_region, Drug_Ingredient, Vital_Signs_Header, Substance_Quantity, Race_Ethnicity, VS_Finding, Injury_or_Poisoning, Medical_History_Header, Alcohol, Triglycerides, Total_Cholesterol, Sexually_Active_or_Sexual_Orientation, Female_Reproductive_Status, Relationship_Status, Drug_BrandName, RelativeTime, Duration, Hypertension, Metastasis, Gender, Oxygen_Therapy, Pulse, Heart_Disease, Modifier, Allergen, Smoking, Substance, Cancer_Modifier, Fetus_NewBorn, Height